\def\year{2021}\relax
\documentclass[letterpaper]{article} % DO NOT CHANGE THIS
\usepackage{aaai21}  % DO NOT CHANGE THIS
\usepackage{times}  % DO NOT CHANGE THIS
\usepackage{helvet} % DO NOT CHANGE THIS
\usepackage{courier}  % DO NOT CHANGE THIS
\usepackage[hyphens]{url}  % DO NOT CHANGE THIS
\usepackage{graphicx} % DO NOT CHANGE THIS
\usepackage{natbib}  % DO NOT CHANGE THIS AND DO NOT ADD ANY OPTIONS TO IT
\usepackage{caption} % DO NOT CHANGE THIS AND DO NOT ADD ANY OPTIONS TO IT
\usepackage{times}
\usepackage{epsfig}
\usepackage{graphicx}
\usepackage{amsmath}
\usepackage{amssymb}
\usepackage{subfigure}
\usepackage{tabularx}
\usepackage{color}
\usepackage{multirow}
\usepackage{rotating}
\usepackage{diagbox}
\usepackage{bm}
\usepackage{caption}
\usepackage[switch]{lineno}
\usepackage{lipsum}

\usepackage{enumitem}
\usepackage{mathtools}
\usepackage{booktabs}

\title{Deep Monocular 3D Human Pose Estimation via Cascaded Dimension-Lifting}
\author {
Changgong Zhang  \thanks{equal contribution} \textsuperscript{\rm 1},
Fangneng Zhan \footnotemark[1] \thanks{corresponding author} \textsuperscript{\rm 2},
Yuan Chang  \textsuperscript{\rm 3} \\
}
\affiliations {
    \textsuperscript{\rm 1} DAMO Academy, Alibaba Group, 
    \textsuperscript{\rm 2} Nanyang Technological University \\
    \textsuperscript{\rm 3} Beijing University of Posts and Telecommunications \\
    changgong.zcg@alibaba-inc.com, fnzhan@ntu.edu.sg, changyuan@bupt.edu.cn  \vspace{20pt}}

\begin{document}
\maketitle

\begin{abstract}

The 3D pose estimation from a single image is a challenging problem due to depth ambiguity. One type of the previous methods lifts 2D joints, obtained by resorting to external 2D pose detectors, to the 3D space. However, this type of approaches discards the contextual information of images which are strong cues for 3D pose estimation. Meanwhile, some other methods predict the joints directly from monocular images but adopt a 2.5D output representation $P^{2.5D} = (u,v,z^{r}) $ where both $u$ and $v$ are in the image space but $z^{r}$ in root-relative 3D space. Thus, the ground-truth information (e.g., the depth of root joint from the camera) is normally utilized to transform the 2.5D output to the 3D space, which limits the applicability in practice. In this work, we propose a novel end-to-end framework that not only exploits the contextual information but also produces the output directly in the 3D space via cascaded dimension-lifting. Specifically, we decompose the task of lifting pose from 2D image space to 3D spatial space into several sequential sub-tasks, 1) kinematic skeletons \& individual joints estimation in 2D space, 2) root-relative depth estimation, and 3) lifting to the 3D space, each of which employs direct supervisions and contextual image features to guide the learning process. Extensive experiments show that the proposed framework achieves state-of-the-art performance on two widely used 3D human pose datasets (Human3.6M, MuPoTS-3D).
\end{abstract}

\section{Introduction}
The goal of 3D human pose estimation is to localize semantic keypoints of single or multiple human bodies in 3D space. It is an essential technique for human behavior understanding and human-computer interaction, yet it is challenging due to the lack of depth information and large variations in human poses, appearances, and camera settings. Traditional approaches often use specialized devices under highly controlled environments, such as multi-view capture \cite{amin2013multi}, marker systems \cite{mandery2015kit} and multi-modal sensing \cite{palmero2016multi}, which require a laborious setup process and thus limits the practical applicability.

\begin{figure}[ht]
\centering
%4.5 8.5
\includegraphics[height=5.5cm,width=8.5cm]{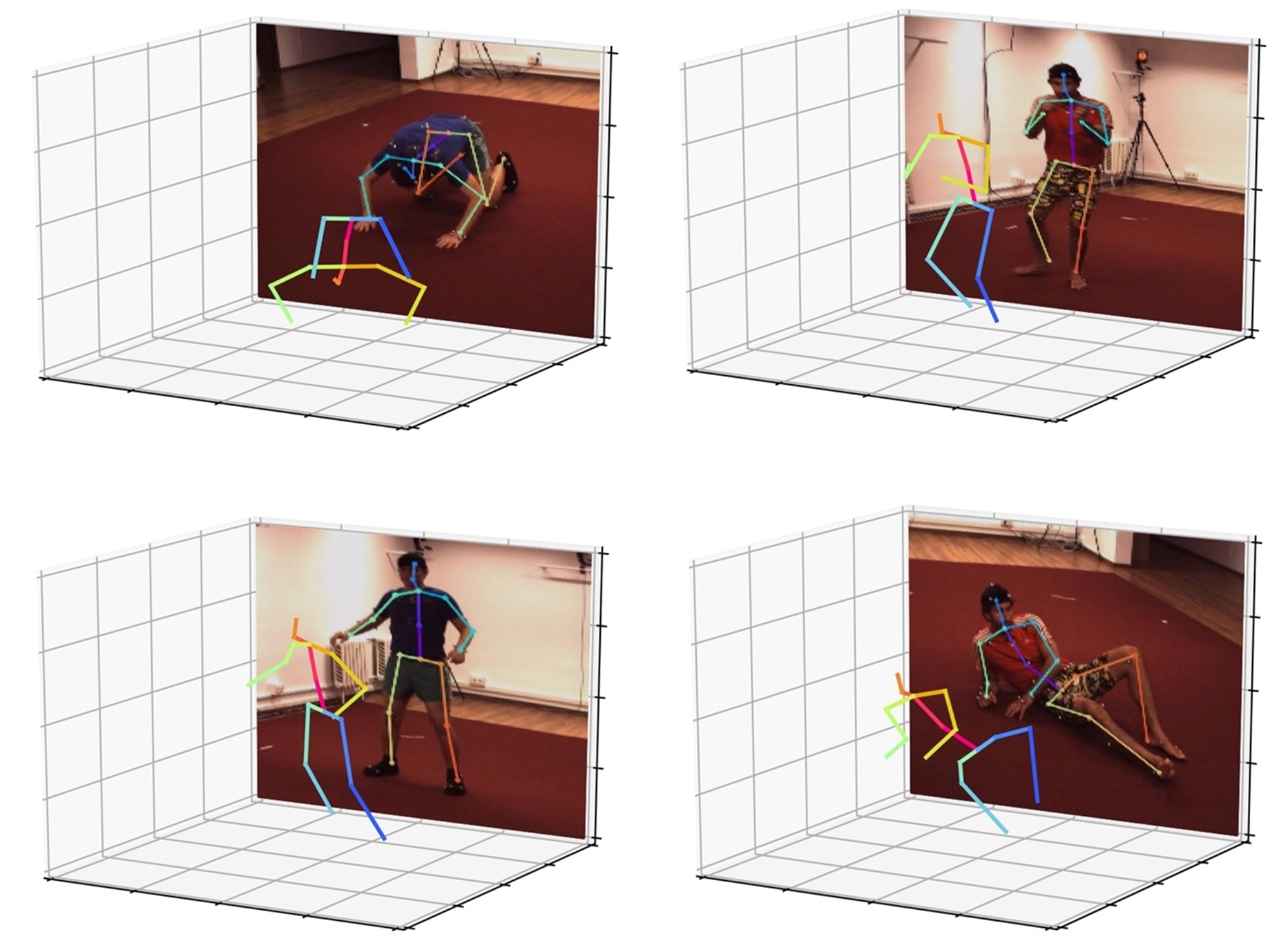} % Reduce the figure size so that it is slightly narrower than the column. Don't use precise values for figure width.This setup will avoid overfull boxes.
\caption{\textbf{Estimated 3D pose by our method }on the Human3.6M dataset. Our approach produces visually correct results even on challenging poses.}
\label{fig1}
\end{figure}

Recently, many learning based methods \cite{martinez2017simple,pavlakos2017coarse,sun2017compositional,sun2018integral,yang20183d} have achieved noticeable performance improvement. As their models take a single cropped image, estimating the absolute camera-centered coordinate of each keypoint is difficult. To handle this issue, many methods estimate the relative 3D pose to a reference point in the body, e.g., the center joint (i.e., pelvis) of a human, called root. Our work also adopts the same strategy as above. 

Several previous researches \cite{moreno20173d,martinez2017simple,wang2019not,sharma2019monocular} take detected 2D keypoints as input and predict corresponding 3D joint locations. This motivation is straightforward but extending them to 3D cases is nontrivial, which has promising results but relies on a high-precision 2D keypoints detector. Still some works \cite{pavlakos2018ordinal,sun2018integral,zhou2019hemlets} adopt a 2.5D pose representation $P^{2.5D} = (u_j,v_j,z_j^{r})_{j \in J} $ where $u_j$ and $v_j$ are the 2D projection of the body joint $j$ and $z_j^{r} = z_{root}-z_j$ represents its metric depth with respect to the root joint. This decomposition of 3D joint locations is superior as additional supervision from in-the-wild images with only 2D pose annotations which can be used for better generalization. While the two dimensions of the output are still in pixel coordinates, the final 3D pose can only be obtained by projection transformation with the help of ground-truth information. 

Our work combines the advantages of the above two pipelines and presents a framework that enables direct estimation of 3D pose from monocular RGB images via dimension-lifting. Generally, the network performs lifting followed by a list from 2D image to 2.5D pose, and then from 2.5D representation to final pose in the physical space.

Firstly, we propose a novel branch called Step-by-Step Heatmap Transformation ($S^3HT$). In this branch, we first adopt a structure regularization methodology for 3D pose estimation, which utilizes skeleton heatmaps to guide keypoints structure learning. The synthetic heatmaps are derived directly from keypoint positions without any additional data annotations. The use of bone information can constrain the search space of keypoints and explore the skeletal relationship. Therefore, the network can predict more accurate 2D keypoint heatmaps. Then we continue to transform the obtained 2D keypoint heatmaps to generate supervision signal in the depth dimension to reduce depth ambiguity. By mapping the image space to a cube, the depth of each point is normalized to the range of the cube's side length. Then we insert the corresponding heatmap into the channel of the cube according to the depth value of each keypoint which plays the role of index. By continuing to learn the volume constructed as described above, the network not only knows the global position relationship but also learns the specific relative depth of each keypoint based on the offset of the index. Some previous methods intend to introduce order ranking for the joint in the depth dimension. However, they only encode three states $(>,<,=)$ for a pair of joints, which destroy the spatial information and cannot reflect different distances in the same state. For example, the states of $+100mm$ and $+1000mm$ are consistent which is not conducive to the subsequent depth prediction. In contrast, our method provides accurate depth distance and more global context which can better alleviate the problem of depth ambiguity in monocular estimation. In a word, we use a step-by-step approach to encode important supervision information in the process of dimension-lifting, from bone heatmap to keypoint heatmap, from keypoint heatmap to the depth order in the cube. 

Then we merge the above feature maps and image context information, use soft-argmax operation to get a set of solid $(u, v, z)$ coordinates where $u,v$ are pixel coordinates. Previous work has shown that 3D pose can be regressed from discrete 2D coordinates and the difficult part is the estimation of $z$ coordinates in 3D space. Compared with the previous methods, the proposed framework can adopts MLP with residual connection to complete the lifting and easily get a high-precision 3D pose with the aid of the accurate predicted depth $z$. Furthermore, the network not only exploits the semantic information but also gets rid of dependence on external 2D keypoint detectors. Figure \ref{fig1} displays qualitative results even on challenging poses. We show that our approach outperforms previous 3D pose estimation methods on several publicly available datasets for 3D single-person and multi-person pose estimation. 
respectively

Overall, our contributions can be summarized as follows.

\begin{quote}
\begin{itemize}
\item We present an end-to-end pipeline directly predicting the 3D pose from monocular images via sequential dimension-lifting, which takes the advantages of 2.5D representation \& 3D lifting paradigms and efficiently guides the learning process. 

\item We propose a novel branch ($S^3HT$), which serially combines multiple heatmap representations to supervise the task in various dimensions. We constrain the search space and increase robust human prior information through 2D bone heatmaps. A new depth representation which effectively reflects the relative depth and differences is also introduced to make the final 3D lifting more accurate.

\item Our method can be applied not only to 3d single-person pose estimation but also 3D multi-person pose estimation with a top-down pipeline. We show that our method significantly outperforms previous methods on both single and multi-person publicly available datasets. 

\end{itemize}
\end{quote}

\section{Related Work}

Compared to 2D human pose estimation, 3D human pose estimation is more challenging since it needs to predict the depth information of body joints. In addition, the training data for 3D human pose estimation are not easy to obtain as 2D human pose estimation. Most existing datasets are obtained under constrained environments with limited generalize ability. 

\begin{figure*}[htbp]
\centering
\includegraphics[scale = 0.95]{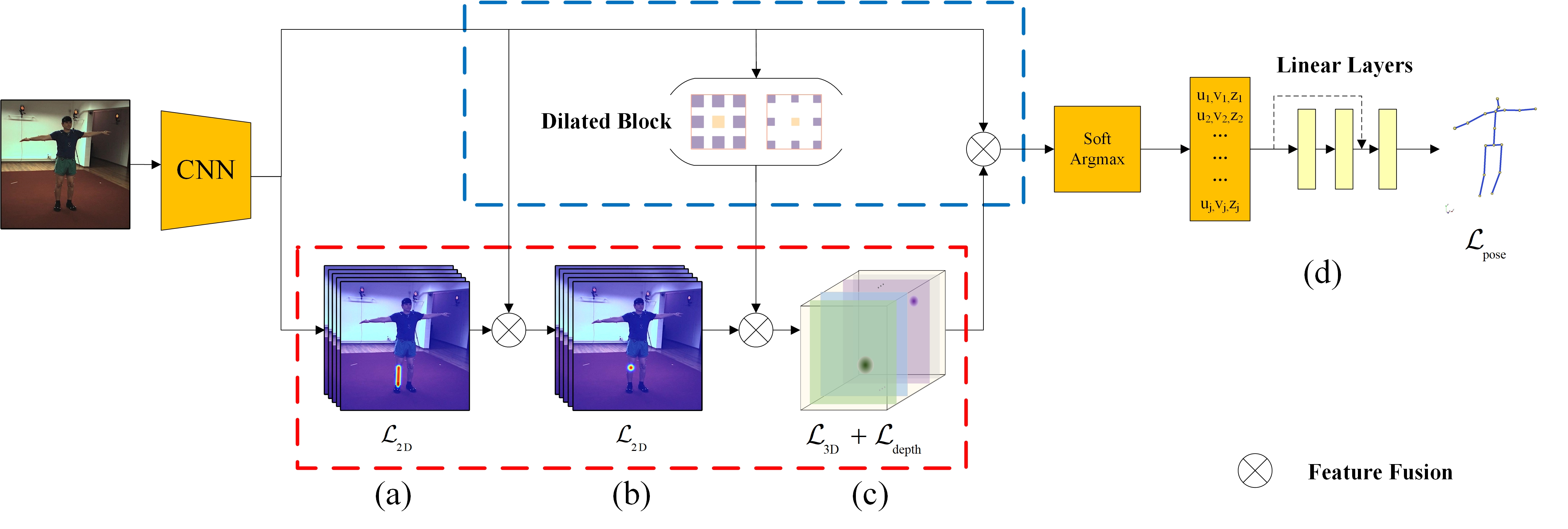}
\caption{The network architecture of our proposed approach. Blue: Original image context branch; Red: Step-by-step heatmap transformation branch ($S^3HT$). (a, b) Bone-Joint heatmap block for skeletal structure learning. (c) The novel Depth-Centric block we exploit. (d) 3D lifting block for the final output.}
\label{fig2}
\end{figure*}

\subsection{3D single-person pose estimation}
Vision-based estimation is one of the most fundamental and challenging problems in computer vision \cite{sun2018integral,habibie2019wild,li20143d,pavlakos2017coarse,pavlakos2018ordinal,zhan2020emlight,zhan2021gmlight,zhan2019esir,zhan2020sagan,zhan2018verisimilar,zhan2019scene,zhan2019gadan,zhan2020aicnet,zhan2019sfgan,zhan2021unite}.
Specially, there are several approaches developed for the estimation of single-person pose. 1) directly map an image to 3D pose, and 2) lift 2D pose to 3D pose.

\cite{zhou2019hemlets,moon2019camera,li2020geometry,li2020cascaded} are based on the first approach. \cite{li20143d} employed a shallow network to regress 3D joint coordinates directly with synchronous task of body part detection with sliding windows. \cite{pavlakos2017coarse} proposed a volumetric representation for 3D human pose and employed a coarse-to fine prediction scheme to refine predictions with a multi-stage structure. \cite{pavlakos2018ordinal} trained the network with additional ordinal depths of human joints as constraints. \cite{sun2018integral} used soft-argmax operation to obtain the 3D coordinates of body joints in a differentiable manner. 
In addition, 
% \cite{habibie2019wild} and \cite{zhou2019hemlets} both utilized intermediate 3D representations and 2D counterparts to regress 3D poses. \cite{moon2019camera} proposed a learning-based, camera distance-aware top-down pipeline for 3D pose estimation from a single RGB image. 
% \cite{li2020geometry} took advantage of the geometric prior for self-supervision. \cite{li2020cascaded} improved the network through a novel data evolution strategy. However, many methods \cite{pavlakos2018ordinal,sun2018integral,zhou2019hemlets} output a 2.5D pose representation in which $(u,v)$ are pixel coordinates. 

\cite{martinez2017simple,zhou2017towards,fang2017learning,yang20183d,sharma2019monocular,pavllo20193d} are based on the second approach. This approach utilizes the high accuracy of 2D human pose estimation network and images from 2D human datasets. They initially localize body keypoints in a 2D space and focus on lifting them to a 3D space. 
% \cite{martinez2017simple} designed a 2D-to-3D pose predictor with only two linear layers. \cite{zhou2017towards} presented a depth regression module to predict 3D pose from 2D heatmaps with a proposed geometric constraint loss for 2D data. 
\cite{yang20183d} utilized adversarial loss to handle the 3D human pose estimation in the wild. Fang et al. \cite{fang2017learning} proposed a pose grammar and considered prior knowledge to incorporate high-level dependencies and relations of human body. \cite{sharma2019monocular} employed a Deep Conditional Variational Autoencoder (CVAE for short) that synthesizes diverse anatomically plausible 3D-pose samples conditioned on the estimated 2D-pose. \cite{pavllo20193d} effectively estimated the 3D pose with a fully convolutional model based on dilated temporal convolutions over 2D keypoints.

\subsection{3D multi-person pose estimation}
The achievements of monocular 3D multi-person pose estimation are based on 3D single person pose estimation and other deep learning methods. \cite{mehta2018single} proposed a bottom-up method by using 2D pose and part affinity fields to infer person instances. An occlusion-robust pose-maps (ORPM) was proposed to provide multi-style occlusion information regardless of the number of people. \cite{rogez2017lcr} proposed a Localization-Classification-Regression Network (LCR-Net) following three stage processing. The localization part detected a human from an input image, and the classification part classified the detected human into several anchor poses. \cite{zanfir2018monocular} proposed a framework relying on detailed semantic representations at both model and image level to guide a combined optimization with feed forward and feed backward stages for 3D multi-person pose and shape estimation. \cite{mehta2020xnect} operated
in subsequent stages and contributed a new architecture called SelecSLS Net. The first stage estimated 2D and 3D pose features along with identity assignments for all visible joints. Then the second stage turned the possibly partial pose features into a complete 3D pose. The third stage applied space-time skeletal model fitting and enforce temporal coherence.

\subsection{Ordinal relations}

In the field of computer vision, ordinal relations have been applied to estimate depth \cite{zoran2015learning,chen2016single} and reflectance \cite{zhou2015learning}. \cite{zoran2015learning} proposed a framework that infers mid-level visual properties of an image by learning about ordinal relationships. \cite{chen2016single} employed a new algorithm for learning to estimate metric depth using only annotations of relative depth. \cite{pavlakos2018ordinal} used a weaker supervision signal provided by the ordinal depths of human joints for 3D human pose. \cite{ronchi2018s} proposed a 3D human pose estimation algorithm that only required relative estimates of depth at training time. \cite{wang2018drpose3d} designed a Pairwise Ranking Convolutional Neural Network (PRCNN) to extract depth rankings of human joints from images. \cite{sharma2019monocular} proposed deep conditional variational autoencoder based model and derived joint-ordinal depth relations from an RGB image and employed them to score and weight-average the candidate 3D-poses.

\section{Proposed Approach}

%In this Section, we describe the proposed approach. Sec.3.1\ref{sec1} discusses the overall network architecture. Sec.3.2\ref{sec2} describes the Bone-Joint Heatmap Block extracting the semantic context of keypoints. In Sec.3.3\ref{sec3}, we describe our novel Depth-Centric block in detail. Finally, Sec.3.4\ref{sec4} predicts the final 3D pose from the previous flow.

\subsection{Network Design}
\label{sec1}
Figure \ref{fig2} depicts the overall architecture of our neural network. It takes a cropped single image as the input and outputs the estimated 3D pose for the target image. A ResNet-50 \cite{he2016deep} backbone is adopted for basic feature extraction for a fair comparison. The framework involves two types of processing modules: original image context branch (indicated by the blue rectangle) and step-by-step heatmap transformation branch ($S^3HT$, indicated by the red rectangle). The $S^3HT$ branch can be further categorized as Bone-Joint heatmap block and Depth-Centric block. 

Firstly, the Bone-Joint heatmap block uses deconvolution to upsample the featuremaps extracted by the backbone and then outputs the 2D skeleton heatmaps through the 2D convolution operation. After fusing with the original image context branch and several 2D convolution operations, high-precision 2D keypoint heatmaps can be predicted. Then the Depth-Centric block takes in joint heatmaps, combined with the original image semantic information which using dilated convolution expanding its receptive field, learning the depth order and distance in the space. Finally the $S^3HT$ branch and original image context branch are joined to predict a three-dimensional featuremap for each joint. 

We perform a soft-argmax operation \cite{sun2018integral} to aggregate information to obtain the 2.5D joint estimation where $(u,v)$ are pixel coordinates. Then the subsequent layer takes the concatenated coordinates and applies a fully connected layer with 1024 output channels. Then it is followed by five blocks that are surrounded by residual connections. For each block, two fully connected layers (1024 channels) followed by Batch Normalization, rectified linear units, and dropout, are stacked for efficiently mapping the pose to high-level features. Finally, the features extracted by the last residual block are fed into an extra linear layer ($N\times3$ channels) to output 3D poses. 

By decomposing the task step by step, we guide the network to perform dimension-lifting from 2D image space to 2.5D representation to 3D spatial space and achieve outstanding performance in the end.

\subsection{Bone-Joint Heatmap Block}
\label{sec2}
\begin{figure}[htbp]
\centering
\includegraphics[scale=0.2]{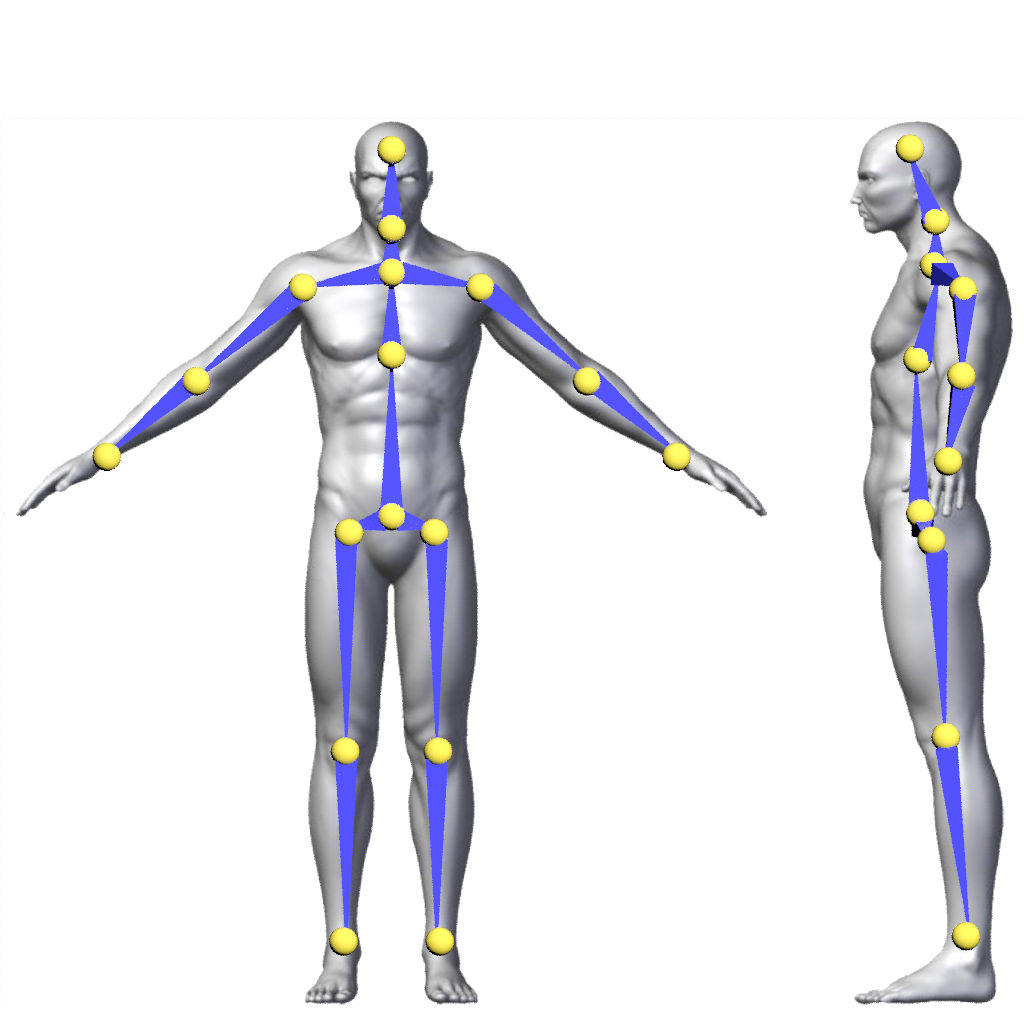} % Reduce the figure size so that it is slightly narrower than the column. Don't use precise values for figure width.This setup will avoid overfull boxes.
\caption{Bones definition and connection relationship.}
\label{fig3}
\end{figure}

Bone-Joint heatmap block learns the skeleton’s structure and keypoint heatmaps in a serial flow. This allows the network to utilize the composite structure representation to learn robust pose information. The same as many pose estimation methods, we adopt heatmap representation to predict accurate 2D coordinates. A ground-truth heatmap consist of a 2D gaussian (with standard deviation of 2 px) centered on the joint location. In order to further introduce the prior information of bones’ length and direction, Gaussian is performed on their connected straight lines to generate a bone heatmap according to the parent-children joint pairs. The Bones definition and connection relationship is shown in Figure \ref{fig3}. Each pixel belongs to $H_{Bone}$ with a Gaussian-alike confidence value as defined bellow.

\begin{equation}
H_{Bone}(p \mid H)=\exp \left(-\frac{Dist(p, \overline{p_{i} p_{j}})}{2 \sigma^{2}}\right)
\end{equation}
where $Dist$ indicates the distance from point to the line of bone.

We adopt the unified representation form of heatmap and can retain large amounts of spatial information. The Bone-Joint heatmap block not only constrains the search space of joints but also promotes the interrelationship between joints by learning the bone structure. Since the learning process is always from coarse to fine, we dynamically reduce the Gaussian kernel size of the bone heatmaps during the training according to the epoch to accelerate the network convergence. It is worth mentioning that the synthetic heatmaps are derived directly from keypoint positions requiring no extra data annotations.

For joint and bone heatmaps, we use mean squared error as the loss functions, where $N$ denotes the number of joints.
\begin{equation}
\mathcal{L}_{2 \mathrm{D}}=\sum_{n=1}^{N}\left\|\mathbf{H}_{n}^{\mathrm{gt}}-\hat{\mathbf{H}}_{n}\right\|_{2}^{2}
\end{equation}

\subsection{Depth-Centric Block}
\label{sec3}

To further bridge the gap between the 2D image and the target 3D human pose, it is necessary to employ immediate supervision for depth. Previous work \cite{pavlakos2018ordinal} adopted the joint ranking loss to encourage the correct pairwise depth ordering in order to alleviate pose ambiguities. However, it only encodes three ordinal states as below rather than the quantitative depth differences. Furthermore, it also tends to lose some important features encoded in the spatial domain because of discrete expression, which would be worse in the pipeline based on the heatmap.

\begin{itemize}[leftmargin = 50pt]
\item +1, if joint i is closer than j,
\item -1, if joint j is closer than i,
\item 0, if their depths are roughly the same 
\end{itemize}

Therefore we propose the Depth-Centric block which still employs the heatmap representation but can provide depth distances and more global position information. Each joint is mapped to a specific location based on its corresponding root-relative depth value and we then arrange the 2D heatmaps of all the joints into a 3D heatmap. Specifically, we normalize the image space to a cube, of which the side length is denoted as $C$. The depth of each joint relative to the root joint is scaled to the range of $[-\frac {C} 2, \frac C 2]$, and we can get the discrete depth value in the range of $[0,C]$ after the depth discretization. We construct 3D featuremap ground-truth equivalent to the cube size and insert the corresponding heatmap into the channel according to the discrete depth value of each keypoint which plays the role of index. In our network, $C$ is set to 64. Note that when several joints have exactly the same discretize values, we also allow multiple joints in the same channel.

By continuing to learn the volume constructed as described above shown in Figure 2(c), the network not only knows the global position relationship but also learns the depth value of each keypoint. We effectively use the expression passed by the previous module without destroying the information of the heatmaps in the spatial domain, and generate a more effective representation for the subsequent learning. It is worth mentioning that this block efficiently uses the semantic information of the original image. A dilated convolution block is employed to extract features from the original image context branch to obtain multiple large receptive fields, which can provide long-range dependencies and capture more global features.

We train the block by minimizing the L2 distance between the estimated and constructed ground-truth. The loss function $\mathcal{L}_{3\mathrm{D}}$ is defined as follows:

\begin{equation}
\mathcal{L}_{3\mathrm{D}}=\sum_{c=1}^{C}\left\|\mathbf{H}_{c}^{\mathrm{gt}}-\hat{\mathbf{H}}_{c}\right\|_{2}^{2}
\end{equation}

But the mean squared error may be the same when the joints are in the wrong channels. For example, if the true label for a given joint is located at the 12th channel, it produces the same error no mater the predicted joint heatmap is located in the 14th or 30th channel, thus it fails to reflect the distance from the ground-truth. This may affects the network convergence speed. Thus we add a conv layer with kernel size of $1\times1$ in parallel to use a new depth loss for assisting. After obtaining the discrete depth values, it is straightforward to turn the problem into a multi-class classification task and adopts ordinal softmax regression loss. The $\mathcal{L}_{\mathrm{depth}}$ formula is as follows and $d_(w, h)$ indicates the depth of the pixel. Minimizing $\mathcal{L}_{\mathrm{depth}}$ ensures that predictions farther from the true label incur a greater penalty than those closer to the true label.

\begin{equation}
\begin{aligned}
\mathcal{L}_{\mathrm{depth}}=\sum_{w}^{W} \sum_{h}^{H} \sum_{c=1}^{d_{(w, h)}} \log \left(\mathcal{P}_{(w, h)}^{c}\right)\\ 
+\sum_{c=d_{(w, h)}+1}^{C}\log\left(1-\mathcal{P}_{(w, h)}^{c}\right)
\end{aligned}
\end{equation}

\begin{equation}
\mathcal{P}_{(w, h)}^{c}=P\left(d_{(w, h)}>c \right)
\end{equation}

\subsection{3D Lifting block}
\label{sec4}
In this section, our goal is to estimate body joint locations in camera space. Formally, our input is a series of points $x \in$ 2.5D, and our output is a series of points $y$ in 3D space. We aim to learn a function that minimizes the prediction error over a dataset of N poses.

\begin{equation}
f^{*}=\min _{f} \frac{1}{N} \sum_{i=1}^{N} \mathcal{L}\left(f\left(\mathbf{x}_{i}\right)-\mathbf{y}_{i}\right)
\end{equation}

There have been some previous works showing that “lifting” 2D joint locations to 3D space that can be solved with a low error rate. But the mapping from 2D coordinates to 3D joints is highly-nonlinear and the discrete joints are not enough to support the high precision depth returned to the physical space. Thus some approaches constrain the solution space by dealing with temporal information. In contrast, we get accurate depth value in the previous network, which can result in a lower-dimensional solution space. So we can get remarkable low error by regression through MLP layers from a single frame.

To avoid quantization errors and allow end-to-end learning, we follow Sun et al. using differentiable soft-argmax regression instead of argmax to get the 2.5D coordinates, the outputs are given as:

\begin{equation}
\mathbf{J}_{n}=\int_{\mathbf{p} \in \Omega} \mathbf{p} \cdot \tilde{\mathbf{H}}_{n}(\mathbf{p})
\end{equation}

\begin{equation}
\tilde{\mathbf{H}}_{n}(\mathbf{p})=\frac{e^{\mathbf{H}_{n}(\mathbf{p})}}{\int_{\mathbf{q} \in \Omega} e^{\mathbf{H}_{n}(\mathbf{q})}}
\end{equation}

And in the final 3D lifting block, mean absolute error is used for minimizing the distance between the estimated and ground-truth 3D pose.

\begin{equation}
\mathcal{L}_{p o s e}=\frac{1}{N} \sum_{n=1}^{N}\left\|\mathbf{J}_{n}^{gt}-\tilde{\mathbf{J}_{n}}\right\|_{1}
\end{equation}

%In order to predict accurate 3D coordinates, we also tried two additional auxiliary methods. The first is fusing the original image-level semantic information of the previous layer to assist the lifting operation. However, the extracted feature is about the cropped image, and a lot of reference information is lost, which will aggravate the depth ambiguity and will not bring gains to the lifting.

We observe that a child and an adult have the same size in the cropped image. However, the child is closer to the camera than the adult. We use a standard bone length to normalize the joints to a uniform height. Specifically, we define the distance from neck to knee as geodesic distance, which accumulates the length of each passing bone. We calculate the average geodesic distance (1077 mm) of the Human3.6M training set and then calculate the scale for each sample and multiply the keypoints by this scale for normalization.

\begin{equation}
{\mathbf{J}}_{3D}^i={\tilde{\mathbf{J}}_{3D}^i}\cdot\frac{1077}{Geo(\tilde{\mathbf{J}}_{3D}^i)}
\end{equation}
where $ \tilde{\mathbf{J}}_{3D}$ means the original 3D joints ground-truth and $Geo(\tilde{\mathbf{J}}_{3D}^i)$ indicates the geodesic distance of the sample $i$.

We add an extra branch to regress the target scale for inference. Although the overall performance can be slightly improved, we later find that only several values of the geodesic distance in the Human3.6M training set because of actors, and the geodesic distances in the test set are very close to the training average. Therefore, the gain may be caused by overfitting, and we finally give up the above strategy.

\section{Experiments}

\begin{table*}[ht]
  \centering
  \small{
  \setlength{\tabcolsep}{1mm}{
    \begin{tabular}{l|cccccccccccccccc}
\toprule 
    Methods & Dir. & Dis. & Eat & Gre. & Phon. & Pose & Pur. & Sit & SitD. & Smo. & Phot. & Wait & Walk & WalkD. & WalkP. & Avg \\
  \midrule
  Chen et al. CVPR'17 & 89.9 & 97.6 & 90.0 & 107.9 &107.3 & 93.6 & 136.1 & 133.1 & 240.1 & 106.7 & 139.2 & 106.2 & 87.0 & 114.1 & 90.6 & 114.2 \\
  Tome et al. CVPR'17& 65.0 & 73.5 & 76.8 & 86.4 & 86.3 & 68.9 & 74.8 & 110.2 &173.9 & 85.0 & 110.7 & 85.8 & 71.4 & 86.3 & 73.1 & 88.4 \\
  \citeauthor{mehta2017monocular} 3DV'17& 57.5 & 68.6 & 59.6 & 67.3 & 78.1 & 56.9 & 69.1 & 98.0 & 117.5 & 69.5 & 82.4 & 68.0 & 55.3 & 76.5 & 61.4 & 72.9  \\
 \citeauthor{martinez2017simple} ICCV'17 & 51.8 & 56.2 & 58.1 & 59.0 & 69.5 & 55.2 & 58.1 & 74.0 & 94.6 & 62.3 & 78.4 & 59.1 & 49.5 & 65.1 & 52.4 & 62.9  \\ 
  \citeauthor{zhou2018monocap} TPAMI'18& 68.7 & 74.8 & 67.8 & 76.4 & 76.3 & 84.0 & 70.2 & 88.0 & 113.8 & 78.0 & 98.4 & 90.1 & 62.6 & 75.1 & 73.6 & 79.9 \\
 \citeauthor{fang2017learning} AAAI'18& 50.1 & 54.3 & 57.0 & 57.1 & 66.6 & 53.4 & 55.7 & 72.8 & 88.6 & 60.3 & 73.3 & 57.7 & 47.5 & 62.7 & 50.6 & 60.4  \\
  \citeauthor{zhao2019semantic} CVPR'19& 47.3 & 60.7 & 51.4 & 60.5 & 61.1 & 47.3 & 68.1 & 86.2 & 55.0 & 67.8 & 49.9 & 61.0 & 60.6 & 42.1 & 45.3 & 57.6  \\
  \citeauthor{li2019generating} CVPR'19& 43.8 & 48.6 & 49.1 & 49.8 & 57.6 & 45.9 & 48.3 & 62.0 & 73.4 & 54.8 & 61.5 & 50.6 & 43.4 & 56.0 & 45.5 & 52.7  \\
  \citeauthor{sharma2019monocular} ICCV'19& 48.6 & 54.5 & 54.2 & 55.7 & 62.6 & 50.5 & 54.3 & 70.0 & 78.3 & 58.1 & 72.0 & 55.4 & 45.2 & 61.4 & 49.7 & 58.0  \\
  Moon et al. ICCV'19& 51.5 & 56.8 & 51.2 & 52.2 & 55.2 & 47.7 & 50.9 & 63.3 & 69.9 & 54.2 & 57.4 & 50.4 & 42.5 & 57.5 & 47.7 & 54.4  \\
  \citeauthor{mehta2020xnect} TOG'20 & - & - & - & - & - & - & - & - & - & - & - & - & - & - & - & 63.6 \\
  \citeauthor{liusemi} IJCAI'20 & 46.5 & 55.0 & 54.7 & 60.2 & 63.3 & 50.2 & 64.2 & 54.0 & 76.0 & 63.2 & 48.9 & 55.5 & 59.3 & 47.6 & 44.0 & 56.2 \\
  \citeauthor{li2020cascaded} CVPR'20 & - & - & - & - & - & - & - & - & - & - & - & - & - & - & - & 50.9 \\
  \midrule
  \textbf{Ours} & \textbf{43.2} & \textbf{50.5} & \textbf{42.7} & \textbf{47.1} & \textbf{50.7} & \textbf{41.1} & \textbf{45.2} & \textbf{58.9} & \textbf{61.6} & \textbf{48.8} & \textbf{60.1} & \textbf{46.3} & \textbf{35.5} & \textbf{51.4} & \textbf{41.0} & \textbf{48.7} \\
    \bottomrule
    \end{tabular}}}%
  \caption{Quantitative comparisons of the mean per-joint position error (MPJPE) on Human3.6M under Protocol \#1.}
  \label{tab:first}%
\end{table*}%

\begin{table*}[ht]
  \centering
  \small{
  \setlength{\tabcolsep}{1mm}{
    \begin{tabular}{l|cccccccccccccccc}
\toprule 
    Methods & Dir. & Dis. & Eat & Gre. & Phon. & Pose & Pur. & Sit & SitD. & Smo. & Phot. & Wait & Walk & WalkD. & WalkP. & Avg \\
  \midrule
  \citeauthor{yasin2016dual} CVPR'16 & 88.4 & 72.5 & 108.5 & 110.2 & 97.1 & 81.6 & 107.2 & 119.0 & 170.8 & 108.2 & 142.5 & 86.9 & 92.1 & 165.7 & 102.0 & 108.3 \\
  Chen et al. CVPR'17& 71.6 & 66.6 & 74.7 & 79.1 & 70.1 & 67.6 & 89.3 & 90.7 & 195.6 & 83.5 & 93.3 & 71.2 & 55.7 & 85.9 & 62.5 & 82.7 \\
  Moreno et al. CVPR'17 & 67.4 & 63.8 & 87.2 & 73.9 & 71.5 & 69.9 & 65.1 & 71.7 & 98.6 & 81.3 & 93.3 & 74.6 & 76.5& 77.7 & 74.6 & 76.5 \\
  \citeauthor{martinez2017simple} ICCV'17 & 39.5 & 43.2 & 46.4 & 47.0 & 51.0 & 41.4 & 40.6 & 56.5 & 69.4 & 49.2 & 56.0 & 45.0 & 38.0 & 49.5 & 43.1 & 47.7  \\
  \citeauthor{sun2017compositional} ICCV'17 & 42.1 & 44.3 & 45.0 & 45.4 & 51.5 & 43.2 & 41.3 & 59.3 & 73.3 & 51.0 & 53.0 & 44.0 & 38.3 & 48.0 & 44.8 & 48.3  \\ 
  \citeauthor{zhou2018monocap} TPAMI'18 & 47.9 & 48.8 & 52.7 & 55.0 & 56.8 & 49.0 & 45.5 & 60.8 & 81.1 & 53.7 & 65.5 & 51.6 & 50.4 & 54.8 & 55.9 & 55.3  \\
  \citeauthor{fang2017learning} AAAI'18& 38.2 & 41.7 & 43.7 & 44.9 & 48.5 & 40.2 & 38.2 & 54.5 & 64.4 & 47.2 & 55.3 & 44.3 & 36.7 & 47.3 & 41.7 & 45.7  \\
  \citeauthor{sun2018integral} ECCV'18& 36.9 & 36.2 & 40.6 & 40.4 & 41.9 & 34.9 & 35.7 & 50.1 & 59.4 & 40.4 & 44.9 & 39.0 & 30.8 & 39.8 & 36.7 & 40.6 \\
  Rogez et al. TPAMI'19& - & - & - & - & - & - & - & - & - & - & - & - & - & - & - & 42.7 \\
  Moon et al. ICCV'19& 32.5 & 31.5 & 41.5 & 36.7 & 36.3 & 31.9 & 33.2 & 36.5 & 44.4 & 36.7 & 38.7 & 31.2 & 25.6 & 37.1 & 30.5 & 35.2 \\
  \midrule
  \textbf{Ours} & \textbf{27.3} & \textbf{29.2} & \textbf{37.2} & \textbf{31.5} & \textbf{33.1} & \textbf{29.2} & \textbf{30.7} & \textbf{32.5} & \textbf{42.0} & \textbf{34.6} & \textbf{36.9} & \textbf{28.0} & \textbf{23.3} & \textbf{34.7} & \textbf{27.5} & \textbf{31.8} \\
    \bottomrule
    \end{tabular}}}%
  \caption{PA MPJPE comparison with state-of-the-art methods on the Human3.6M dataset using Protocol \#2. The results of all approaches are obtained from the \cite{moon2019camera}.}
  \label{tab:second}%
\end{table*}% 

\subsection{Dataset and Evaluation Metric}
\subsubsection{Human3.6M dataset.}
Human3.6M \cite{ionescu2013human3} is the largest 3D human pose estimation benchmark with accurate 3D labels. It contains 3.6 million RGB images captured by a MoCap System in an indoor environment, in which 7 professional actors were performing 15 activities from 4 camera viewpoints such as walking, eating, sitting, making a phone call and engaging in a discussion, etc. 2D joint locations and 3D ground truth positions are available, as well as projection (camera) parameters and body proportions for all the actors. Two evaluation metrics are widely used. The first one computes the mean Euclidean distance for all the joints after aligning the root joints (i.e. pelvis) between the predicted and ground-truth poses, referred as MPJPE. The second one is MPJPE after applying an additional rigid transformation (i.e., Procrustes analysis (PA) ) to the predicted pose as an enhancement. This metric is called PA MPJPE. 

\subsubsection{MuCo-3DHP and MuPoTS-3D datasets.}
These are the 3D multi-person pose estimation datasets proposed by Mehta et al. The training set, MuCo-3DHP, is generated by compositing the existing MPI-INF-3DHP 3D single-person pose estimation dataset \cite{mehta2017monocular}. The test set, MuPoTS-3D dataset, was captured at outdoors and it includes 20 real-world scenes with ground-truth 3D poses for up to three subjects. The ground-truth is obtained with a multi-view marker-less motion capture system. For evaluation, a 3D percentage of correct keypoints (3DPCK) is used after root alignment with ground-truth. It treats a joint’s prediction as correct if it lies within a 15cm from the ground-truth joint location.

\subsection{Experimental Protocol}

\subsubsection{Human3.6M dataset.}
Two experimental protocols are widely used. Protocol 1 uses five subjects (S1, S5, S6, S7, S8) in training and two subjects (S9, S11) in testing. MPJPE is used as an evaluation metric. Protocol 2 uses six subjects (S1, S5, S6, S7, S8, S9) in training and S11 in testing following \cite{zhou2018monocap,rogez2019lcr,moon2019camera}. PA MPJPE is used as an evaluation metric. We use every 5th and 64th frames in videos for training and testing following \cite{sun2017compositional,sun2018integral}. When training, besides the Human3.6M dataset, we used additional MPII 2D human pose estimation dataset \cite{andriluka20142d} following \cite{pavlakos2017coarse,sun2017compositional,sun2018integral,moon2019camera}. Each mini-batch consists of half Human3.6M and half MPII data. For MPII data, the loss value of the z-axis becomes zero. 

\subsubsection{MuCo-3DHP and MuPoTS-3D datasets.}
Following the previous works \cite{mehta2018single,moon2019camera}, we use images from the COCO dataset \cite{lin2014microsoft} except for images with humans for augmentation and an additional COCO 2D human keypoint detection dataset when training our models on the MuCo-3DHP dataset. Each mini-batch consists of half MuCo-3DHP and half COCO data. For COCO data, loss value of z-axis becomes zero.

\subsection{Implementation Details}

We implement our method in PyTorch. The backbone is initialized with the publicly released ResNet-50 pre-trained on the ImageNet dataset. In order to use both images with 3D annotations (e.g., Human3.6M) and 2D annotations (e.g., MPII), the training procedure contains two stages. Because 3D lifting block can only use 3D ground-truth supervision, we train previous network and 3D lifting block sequentially. For previous network, we use the Adam as the optimizer and train the network for 20 epochs with a batch size of 64 and the initial learning rate is set to 0.001 and reduced by a factor of 10 at the 17th epoch. We use $256\times256$ as the size of the input image and perform data augmentation including rotation ($\pm30^{\circ}$), horizontal flip, synthetic occlusion in training. Then for the 3D lifting block we still follow the above strategy. The only difference is that we just use 3D dataset (e.g., Human3.6M) for fine-tuning. The training is performed on two NVIDIA P100 GPUs.

\subsection{Comparison with State-of-the-art Methods}

\subsubsection{Human3.6M dataset.}
We compare our approach with state-of-the-art techniques under two protocols on the Human3.6M dataset \cite{ionescu2013human3} as shown in Tables \ref{tab:first} and \ref{tab:second}. As can be seen, our method outperforms all competing single-frame methods and obtains average errors of 48.7mm and 31.8mm under two evaluation protocols. In addition, some approaches such as \cite{sun2018integral,zhou2019hemlets} use the camera intrinsics and the ground-truth distance of the subject from the camera to convert their $(u,v)$ predictions to $(x,y)$. Therefore, their reported results are not representative of the 3D pose performance. Our method achieves comparable performance despite not using any ground-truth information in inference time.

\begin{table}[ht]
  \centering
  \small{
  \setlength{\tabcolsep}{4mm}{
    \begin{tabular}{l||c}
\toprule 
    Methods &\qquad 3DPCK(\%) \qquad \qquad \\
  \hline
  \citeauthor{mehta2018single} 3DV'18 & 65.0 \\
  Rogez et al. TPAMI'19 & 68.9 \\
  \citeauthor{dabral2019multi} 3DV'19 & 71.3 \\
  Moon et al. ICCV'19 & 81.8 \\
  \citeauthor{mehta2020xnect} TOG'20 & 72.1 \\
  \citeauthor{benzine2020pandanet} CVPR'20 & 72.0 \\
  \hline
  \textbf{Ours} & \textbf{83.3}\\
    \bottomrule
    \end{tabular}}}%
  \caption{Comparison of our approach on the MuPoTS-3D benchmark dataset. The metric used is 3D percentage of correct keypoints (3DPCK), so higher is better.}
  \label{tab:third}%
\end{table}%

\subsubsection{MuCo-3DHP and MuPoTS-3D datasets.} Our approach can complete 3D multi-person pose estimation with a top-down pipeline. We quantitatively evaluate our method’s performance on the MuPoTS 3D monocular multi-person benchmark dataset from Mehtaet al. We report the 3DPCK (percentage of joint prediction within a 15cm ball centred on ground-truth) per sequence, averaged over the subjects for which ground truth is available. Results are reported in Table \ref{tab:third}. We establish a new state-of-the-art performance of 83.3\% compared to 81.8\% which also belongs to the top-down pipeline.

\subsection{Ablation Study}

To verify the impact and performance of each component in the network, we conduct ablation experiments on the Human3.6M dataset under Protocol \#1. We examine the effectiveness of using different supervision we proposed. We evaluate the model trained without any intermediate supervision (Baseline), with keypoint heatmap supervision only, with Bone-Joint heatmap block supervision, and with Depth-Centric block supervision (Full). All of these design variants are evaluated with the same experimental setting (including training data, network architecture). Our baseline is based on the combination of the PoseNet in \cite{moon2019camera} and the 3D Lifting block.

The detailed results are presented in Table \ref{tab:forth}. Using keypoint heatmaps supervision for training, the prediction error is reduced by 2.1mm compared to the baseline. The Bone-Joint heatmaps supervision provide 1.2mm lower mean error compared to the keypoint heatmaps supervision. By combining all the component, our method achieves significant performance improvements, and the MPJPE decreases from 54.1(mm) to 48.7 (mm). It is worth mentioning that our network only adds a few parameters compared with the previous methods.

\begin{table}[htbp]
  \centering
  \small{
  \setlength{\tabcolsep}{5mm}{
    \begin{tabular}{l|c|c}
\toprule 
    Methods &Supervision& MPJEP(mm) \\
  \hline
  Baseline  & \-- & 54.1\\
  w/ JH  & JH & 52.0\\
  w/ BJ & JH+BH & 50.8\\
  Full  & JH+BH+DC & 48.7\\
    \bottomrule
    \end{tabular}}}%
  \caption{Ablative study on the effects of the intermediate supervision we proposed. The evaluation is performed on Human3.6M under Protocol \#1 with MPJPE(mm). JH: joint heatmap, BH: bone heatmap, BJ:bone-joint block, DC: depth-centric block.}
  \label{tab:forth}%
\end{table}%

\section{Conclusion and Future Work}
This paper presents a novel framework for monocular 3D pose estimation, consisting of a clean CNN architecture and effective intermediate supervisions for network optimization. Our work takes the advantages of 2.5D representation and 2D-to-3D lifting methods, decomposing the task into several sequential sub-tasks, helping network completing dimension-lifting step by step. The proposed system is very helpful to bridge the 2D-3D domain gap in the learning procedure and achieves outstanding performance without using any ground-truth information during inference. In the future, we intend to explore the extension to temporal domain, which can be easily plugged into our framework. Moreover, we will improve the joints depth representation and investigate more prior constraints for the network.

\bibliography{cite}
\end{document}